\pdfoutput=1

\documentclass[11pt]{article}

\usepackage[review]{EMNLP2023}

\usepackage{times}
\usepackage{latexsym}

\usepackage[T1]{fontenc}

\usepackage[utf8]{inputenc}

\usepackage{microtype}

\usepackage{inconsolata}

\usepackage{times}
\usepackage{latexsym}
\usepackage{amsthm,amsmath,amssymb}
\usepackage{mathrsfs}
\usepackage{bm}
\usepackage{booktabs}
\usepackage{color}
\usepackage{graphicx}
\usepackage{multirow}
\usepackage{makecell}

\usepackage{float}
\usepackage{url}
\usepackage{xcolor, soul}
\definecolor{amethyst}{rgb}{0.6, 0.4, 0.8}
\sethlcolor{amethyst}

\usepackage{adjustbox}
\usepackage{fancyvrb}
\usepackage{caption}
\usepackage{subcaption}
\usepackage{pdfpages}

\renewcommand{\bold}{\mathbf}

\newcommand{\squishlist}{
 \begin{list}{$\bullet$}
  { \setlength{\itemsep}{0pt}
     \setlength{\parsep}{3pt}
     \setlength{\topsep}{3pt}
     \setlength{\partopsep}{0pt}
     \setlength{\leftmargin}{1.5em}
     \setlength{\labelwidth}{1em}
     \setlength{\labelsep}{0.5em} } }

\newcommand{\squishend}{
  \end{list}  }

\usepackage[small,compact]{titlesec}

\title{Temporal reasoning for timeline summarisation in social media}

\author{Jiayu Song$^{1}$, Mahmud Elahi Akhter$^{1}$, Dana Atzil-Slonim$^{2}$, Maria Liakata$^{1,3}$ \\
        $^1$Queen Mary University of London, London, UK \\ 
        $^2$Bar-Ilan University, Israel\\ 
        $^3$The Alan Turing Institute, London, UK\\
        \texttt{\{jiayu.song,m.akhter,m.liakata\}@qmul.ac.uk}\\
        \texttt{dana.slonim@gmail.com}\\}

\begin{document}
\maketitle

\begin{abstract}

This paper explores whether enhancing temporal reasoning capabilities in Large Language Models (LLMs) can improve the quality of timeline summarisation, the task of summarising long texts containing sequences of events, such as social media threads. We first introduce \textit{NarrativeReason}, a novel dataset focused on temporal relations among sequential events within narratives, distinguishing it from existing temporal reasoning datasets that primarily address pair-wise event relations. Our approach then combines temporal reasoning with timeline summarisation through a knowledge distillation framework, where we first fine-tune a teacher model on temporal reasoning tasks and then distill this knowledge into a student model while simultaneously training it for the task of timeline summarisation. Experimental results demonstrate that our model achieves superior performance on out-of-domain mental health-related timeline summarisation tasks, which involve long social media threads with repetitions of events and a mix of emotions, highlighting the importance and generalisability of leveraging temporal reasoning to improve timeline summarisation.

\end{abstract}
\section{Introduction}
Timeline summarisation organizes and presents a sequence of events in a coherent and concise manner \cite{steen2019abstractive, Li0YWGJM21, HuMN24}. It involves extracting event-related timelines and then summarising them \cite{HuMN24,crisisltlsum}. Researchers generally create event graphs \cite{Li0YWGJM21} or cluster event related timelines \cite{HuMN24} to identify relevant events.
Recent work~\cite{SongCTIAL24} has introduced the challenging task of social media timeline summarisation, especially in the context of capturing fluctuations in individuals' state of mind as reflected in posts shared online over time. 
In these posts, numerous events may occur without explicit timestamps, requiring contextual inference to determine their chronological sequence. Moreover, mental health-related events are not easy to identify: they can be connected to an individual's emotions, interpersonal interactions, and the entire timeline is necessary to provide enough context \cite{SongCTIAL24}. It is particularly challenging to identify events pertaining to psychological states and to extract these from posts. When generating mental health related summaries from longitudinal posts, models need to understand related events and maintain temporal consistency to make inferences. This raises the question of whether temporal reasoning can be leveraged to enhance the quality of complex timeline summaries. 

Temporal reasoning involves understanding and processing temporal information in text to deduce time-based relations between events \cite{GeorgOverview}. \citet{ZhouKNR19} categorises temporal commonsense reasoning with respect to five aspects (duration, temporal ordering, typical time, frequency and stationarity).
Subsequently, \citet{TanNB23, jainmodels} explore the temporal reasoning capabilities of Large Language Models (LLMs) with respect to temporal commonsense aspects. LLMs with a strong understanding of temporal context can perform better on downstream tasks, including storytelling, natural language inference, timeline comprehension and tracking user status \cite{jainmodels}.
Thus temporal commonsense reasoning is beneficial for timeline summarisation, as it helps maintain temporal consistency and the correct event order \cite{abs2308,vashishtha-etal-2020-temporal}. 
Despite the evidenced connection between temporal reasoning and timeline summarisation, recent work \cite{chan2024exploring, FengZ0JR23, MS0Z024, zhang2024narrative} has primarily focused on improving the temporal reasoning capabilities of LLMs, without exploring how they impact downstream tasks, such as timeline summarisation. 

Here we propose combining temporal reasoning with timeline summarisation using LLMs, to enhance the generation of timeline summaries. Specifically, we first fine-tune a teacher model using a novel temporal reasoning dataset (NarrativeReason) and then distill temporal reasoning knowledge into a smaller student model, which is simultaneously fine-tuned on the timeline summarisation task, trained and tested in separate domains. 

\noindent We make the following contributions:

\squishlist
\item We are the first to explore how enhancing temporal reasoning in LLMs can improve timeline summarisation.

\item Based on the timelines derived in NarrativeTime \cite{rogers2019narrativetime} from the TimeBankNT corpus \cite{CassidyMCB14}, we develop a new dataset for temporal reasoning, \textit{NarrativeReason}. Unlike existing temporal reasoning datasets \cite{TanNB23, chu2023timebench, wang2023tram},  \textit{NarrativeReason} focuses on the temporal relations among a series of events within a story rather than distinct event pairs. 
This can help LLMs process a series of events to generate a coherent and accurate timeline summary.

\item 
We fine-tune a large LLM on \textit{NarrativeReason} and distill its knowledge to a smaller model, which is fine-tuned for the task of timeline summarisation. The resulting fine-tuned smaller LLM is applied to a completely different domain from the one it is trained on. Experimental results show that our model achieves the best performance on the timeline summarisation dataset by \cite{SongCTIAL24}. Not only does it generate more accurate summaries, but it also reduces hallucinations in LLMs.

\item We show why knowledge distillation works well, and how it induces better learned representations, through activation analysis of the fine-tuned smaller LLM.
\squishend



\section{Related Work}

\noindent\textbf{Temporal reasoning for LLMs}
Temporal reasoning in Natural Language Processing (NLP) is the ability to understand and process information related to time within natural language text. It includes reasoning about the chronology and duration of events, and understanding and capturing different temporal relations \cite{vashishtha-etal-2020-temporal}. 
Despite the impressive performance of Large Language Models (LLMs), like GPT-4, across a wide range of tasks (e.g. translation, generation), they have been shown to  perform sub-optimally in temporal reasoning \cite{wang2023tram, chu2023timebench, qiu2023large}. However the ability to perform temporal reasoning is crucial for understanding narratives \cite{nakhimovsky-1987-temporal, jung-etal-2011-building, cheng2013temporal}, answering questions \cite{bruce1972model, khashabi2019reasoning, ning-etal-2020-torque}, and summarising events \cite{jung2011building, vashishtha-etal-2020-temporal}. Consequently, efforts are being made to enhance the temporal reasoning capabilities of LLMs \cite{10163070} \cite{huang2024}.
To increase understanding of temporal expressions, \citet{TanNB23} introduced the TEMPREASON dataset which addresses three types of relations (time-time, time-event, event-event). TEMPREASON was used to fine-tune a LLM to improve its temporal reasoning, and performance of different LLMs on this dataset showed it is challenging for LLMs to capture the temporal relations between different events. \citet{xiong2024large} use an aligned timeline to 
improve an LLM's temporal reasoning by translating the context into a temporal graph,identifying valid time expressions and generating related temporal knowledge. The temporal relations between events are inferred based on specific times (e.g., year of event). However in a narrative, events often occur without a clear indication of time.\\
\noindent\textbf{Temporal reasoning for summarisation}
\citet{jung2011building} developed a natural language understanding (NLU) system with a temporal reasoning component to create comprehensive timelines, applied to medical records,  presenting medical history in a more intuitive way. They found that temporal reasoning in NLU is tightly integrated into the NLP system’s deep semantic analysis and can help a LM analyze temporal relations between different events, which is beneficial for event or news summarisation \cite{vashishtha-etal-2020-temporal}. However, despite the evidenced connection, few studies explore how improvements in temporal reasoning in LLMs directly benefit downstream tasks such as text summarisation.

\vspace{-0.2cm}
\section{Methodology}\label{method}
\vspace{-0.2cm}
\textbf{Task}\label{task}
 Given an individual's timeline (a series of posts between two dates \cite{identifying}), the goal is to generate an abstractive summary that reflects changes in the individual over time ~\cite{SongCTIAL24}. 
\begin{figure*}
\centering
\includegraphics[width=0.7\textwidth]{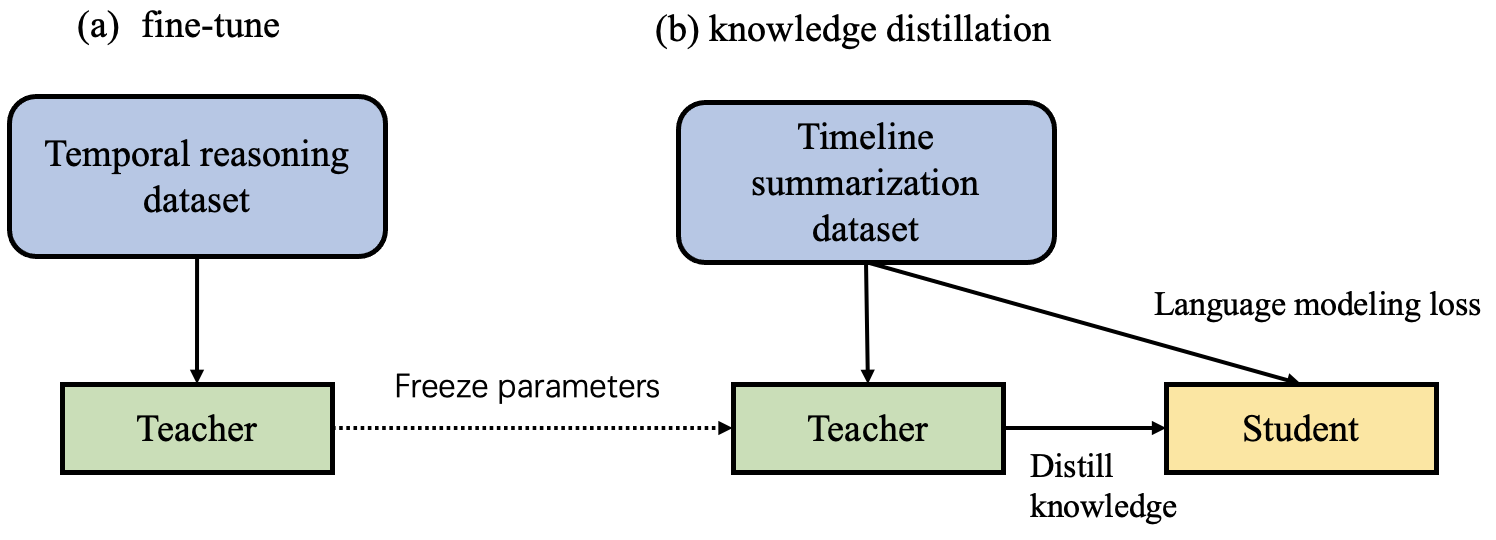}
\caption{Overview of proposed method. (a) represents fine-tuning the teacher model on the temporal reasoning dataset; (b) In the KD process, we input the timeline summarisation dataset into both the Teacher and Student models, transferring temporal reasoning knowledge while using it to assist the timeline summarisation task to fine-tune Student model.}
\label{fig:overview}
\vspace{-0.5cm}
\end{figure*}
\subsection{Proposed architecture}\label{architecture_overview}
To generate timeline summaries on social media we consider two sub-processes (see Fig.~\ref{fig:overview}):\\
(1) Improving temporal reasoning. We fine-tune a large LLM as a teacher model on our `NarrativeReason' dataset \S\ref{method:teacher}.\\
(2) After fine-tuning the teacher model, we freeze its parameters. At this stage, we fine-tune a student model (a smaller LLM) on timeline summarisation from news~\cite{chen2023follow}. During this process, the teacher model transfers temporal reasoning knowledge to the student model, while the student simultaneously leverages the acquired temporal reasoning knowledge to perform timeline summarisation. For knowledge distillation (KD), we adopt three different strategies: Neuron Selectivity Transfer (NST), Contrastive Representation Distillation (CRD) and Probabilistic Knowledge Transfer (PRT).
When generating the timeline summary, we conduct experiments on the TalkLife dataset, which pertains to a very different domain (mental health) to the one the student is trained on (news). We prompt the student model to generate mental health related summaries pertaining to aspects such as diagnostic states, inter- and intra- personal relationships and fluctuations in mood following~\cite{SongCTIAL24}. 

\subsection{Teacher Model and NarrativeReason dataset}\label{method:teacher}
Here the goal is to improve an LLM's temporal reasoning. Evidence has shown that fine-tuning on datasets such as TEMPLAMA may enable an LLM to memorise the most frequent answer rather than develop temporal reasoning~\cite{TanNB23}. In other words, the model does not truly learn the meaning of temporal relations, such as "before" and "after". We hypothesise that this is because most temporal reasoning datasets involve pairs of events rather than multiple events. Processing a sequence of events requires more intricate reasoning, including recognising patterns, dependencies, and causal chains among multiple events. This is useful for more sophisticated tasks such as narrative comprehension and timeline summarisation, where understanding the full sequence of events is crucial.
To prevent the LLM from learning shortcuts and memorising the most frequent answer, we created a temporal reasoning dataset \textit{NarrativeReason}, which contains relations between a series of events based on a given narrative.
\vspace{-0.1cm}

\textbf{Event extraction}
To create NarrativeReason we restructured the NarrativeTime dataset~\cite{rogers2019narrativetime}, which in turn had re-annotated TimeBankDense~\cite{CassidyMCB14} 
with a timeline-based annotation framework, NarrativeTime. \citet{rogers2019narrativetime} have annotated all possible temporal links (TLINKS) between all events occurring within a narrative, thus providing temporal relations for the entire sequence of events (timeline) rather than just between event pairs to get obtain the NarrativeTime dataset. Here, TLINKs usually denote the event order information (e.g., before, after, during). 
Events more broadly and especially in the context of temporal reasoning, are represented as relation triples where the event trigger is the head of the verb phrase \cite{ning-etal-2018-cogcomptime, pustejovsky2003timebank}, linking the corresponding arguments. However, NarrativeTime ~\cite{rogers2019narrativetime} only denotes the type of event, annotated at the level of the verb, e.g. for the sentence "the value of the Indonesian stock market has fallen by twelve percent" this is annotated in the NarrativeTime dataset as "the value of the Indonesian stock market has <EVENT class="OCCURRENCE" eid="e7"> fallen </EVENT> by twelve percent". However, this only constitutes a denser temporal annotation without providing event triples and is therefore unsuited for temporal reasoning training tasks. 

Thus we reconstructed NarrativeReason, to augment it with event triples that can be used for temporal reasoning training.
Specifically, we filtered the NarrativeTime dataset, keeping only annotated verbs to represent events. In order to represent an event, we use verbs as triggers to extract relational triples e.g. \textit{<Indonesian stock market value, fallen, by twelve percent>} now represents the event annotated in the NarrativeTime dataset as \textit{fallen}. Then, we use these triples to construct the temporal relations of a series of events. (e.g. \textit{Event <Indonesian stock market value, fallen, by twelve percent> is BEFORE Event <financial week, turning, bad for Asia>}).

\begin{figure*}
\centering
\includegraphics[width=1\textwidth]{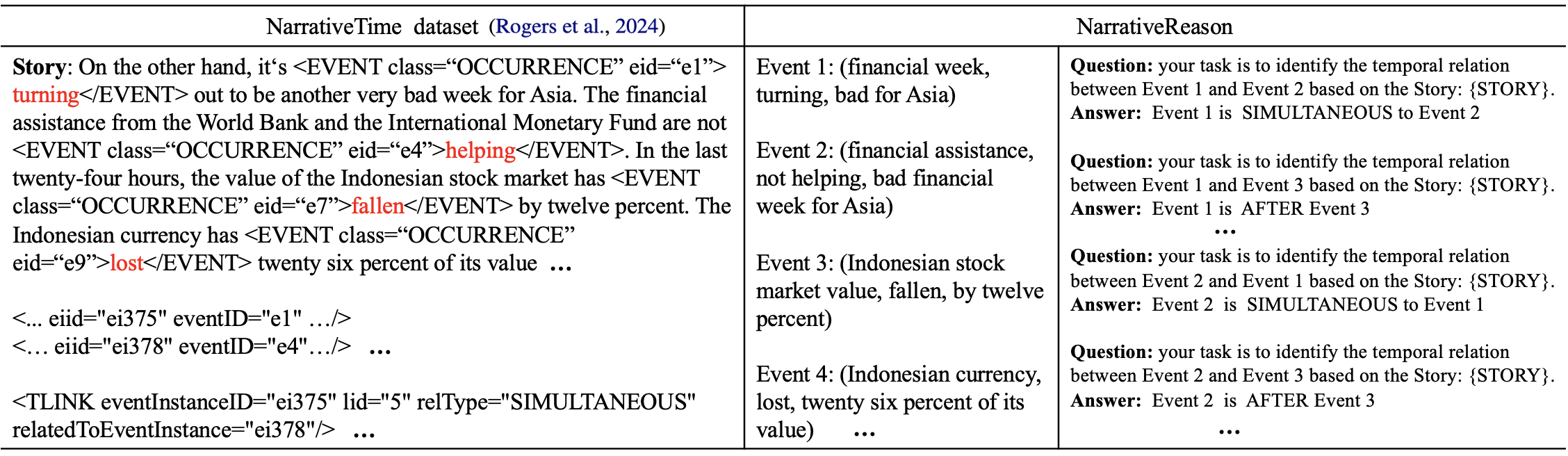}
\caption{The temporal relations between events. The text in the left column comes from the \textit{NarrativeTime dataset}, and we filtered it to keep only verbs. We represent events triggered by verbs using relational triples, as shown in the middle column. In the right column, we construct question/answer pairs for all events.}
\label{fig:temporal}
\vspace{-0.55cm}
\end{figure*}

\textbf{Dataset construction}
For a given narrative, we consider the temporal relations between all events, and construct question/answer pairs for event-event relations, addressing the chronological order of events (‘before’, ‘after’, ‘during’, and ‘simultaneous’ \cite{TanNB23}). Specifically, we obtain the temporal relations of all events and then use \textbf{question answering} prompts to reconstruct the dataset. \textbf{Question}: your task is to identify the temporal relation between \textit{EVENT A} and \textit{EVENT B}: based on the Story: \textit{STORY}. \textbf{Answer}: \textit{EVENT A} temporal relation (BEFORE{/} AFTER{/} INCLUDES{/} IS\_INCLUDED{/} SIMULTANEOUS) \textit{EVENT B} \cite{TanNB23}. Although a single question-answer pair is used to determine the temporal relation between a pair of events, for a complete narrative, we construct multiple question-answer pairs to cover the temporal relations among all events. This ensures that the model is exposed to all temporal relations between all events in the story. Fig.~\ref{fig:temporal} shows the data format of \textit{NarrativeReason}.

\textbf{Fine-tuning task}
We apply supervised fine-tuning (SFT) on a large LLM (teacher model) utilising Low-Rank Adaptation (LoRA) \cite{hu2021lora}. The input and output of the model are the temporal questions and corresponding answers respectively. Our experiments show that indeed fine-tuning on NarrativeReason improves performance on established temporal reasoning tasks (Appendix \ref{appen:teacher_result}) .

\subsection{Student Model}
After we fine-tune a teacher model on the NarrativeReason dataset, we transfer the temporal reasoning knowledge to a student model, while, also fine-tuning the student on a news timeline summarisation dataset \cite{chen2023follow}. Thus we aim for the student to learn temporal reasoning and also use this ability when generating timeline summaries (step (b) in Fig.~\ref{fig:overview}). We fine-tune Phi-3-mini-4k-instruct as a student model. We use three knowledge distillation (KD) objectives to transfer knowledge from the teacher to the student: Neuron Selectivity Transfer (NST) \cite{huang2017like}, transfers heatmap like spatial activation patterns of teacher neurons to student neurons; Contrastive Representation Distillation (CRD) \cite{tian2019contrastive}, maximises the mutual information between the teacher and student representations with contrastive learning; Probabilistic Knowledge Transfer (PRT) \cite{passalis2018learning}, matches the probability distribution of the data in the feature space of teacher and student models. 
\paragraph{PRT:} 
Learning a significantly smaller model that accurately recreates the whole geometry of a complex teacher model is often impossible. \citet{passalis2018learning} uses the conditional probability distribution to describe the samples. Here, $\bold{Y_t}={\{\bold{y_t}}_{1}, {\bold{y_t}}_{2}, ..., {\bold{y_t}}_{l}\}\in\mathbb{R}^{vocab_t}$ denote the output logits of the teacher model, and $\bold{Y_s}={\{\bold{y_s}}_{1}, {\bold{y_s}}_{2}, ..., {\bold{y_s}}_{l}\}\in\mathbb{R}^{vocab_s}$ denote the output logits of the student model, where ${\bold{y_t}}$ and ${\bold{y_s}}$ are vectors and $l$ is sentence length, ${vocab_t}$ and ${vocab_s}$ are the vocabulary sizes of teacher and student models respectively. We can define the conditional probability distribution for the teacher model as Eq.~\ref{teacher_condition}, and student model as Eq.~\ref{student_condition}, where $K$ is a symmetric kernel with scale parameter $\sigma$.
\begin{equation}
    \begin{aligned}
    {p_{i|j}}=\frac{K({\bold{y_t}_i},\bold{y_t}_j;2\sigma_{t}^2)}
    {\sum_{k=1,k\neq{j}}^{l}{K(\bold{y_t}_k,\bold{y_t}_j;2\sigma_{t}^2)}}
    \end{aligned}\label{teacher_condition}
\end{equation}
\vspace{-.25em}
\begin{equation}
    \begin{aligned}    {q_{i|j}}=\frac{K(\bold{y_s}_i,\bold{y_s}_j;2\sigma_{s}^2)}
    {\sum_{k=1,k\neq{j}}^{l}{K(\bold{y_s}_k,\bold{y_s}_j;2\sigma_{s}^2)}}
    \end{aligned}\label{student_condition}
\end{equation}
In equations Eq.~\ref{teacher_condition} and Eq.~\ref{student_condition}, we use the cosine similarity kernel (no $\sigma$), allowing for more robust affinity estimations:
\vspace{-.5em}
\begin{equation*}
    \begin{aligned}
    {K_{cosine}({\bold{y_t}_i}, {\bold{y_t}_j})}=\frac{1}{2}(\frac{{\bold{y_t}_i^\mathrm{T}}\bold{y_t}_j}{\Vert{\bold{y_t}_i}\Vert{_2}\Vert{\bold{y_t}_j}\Vert{_2}}+1) \in[0,1].
    \end{aligned}\label{cosine_kernel}
\end{equation*}
We use Kullback-Leibler (KL) divergence to calculate the distance between the conditional probability distributions of the teacher and student models: 
\vspace{-.75em}
\begin{equation*}
    \begin{aligned}
    {L_{PKT}}={\sum_{i=1}^{l}\sum_{j=1,i\neq{j}}^{l}}
   {p_{i|j}}\log({\frac{p_{i|j}}{q_{i|j}}}).
    \end{aligned}
    \vspace{-.25em}
\end{equation*}
\paragraph{NST}~matches the distributions of neuron selectivity patterns between teacher and student networks. We transfer the last hidden layer $\bold{T}=\bold{t(x)}$ of 
the teacher model to the last hidden layer $\bold{S}=\bold{s(x)}$ of the student model given input text $\bold{x}$. Specifically, we transfer neuron selectivity knowledge from $\{{\bold{t(x)}}_{*,i}\}_{i=1}^{N}$ to $\{{\bold{s(x)}}_{*,i}\}_{i=1}^{M}$, where $N$ and $M$ are the hidden state dimensions. Then we follow  \citet{huang2017like} in using Maximum Mean Discrepancy (MMD) to calculate the distance between the activation patterns of student $\{{\bold{s(x)}}_{*,i}\}_{i=1}^{M}$ and teacher neurons $\{{\bold{t(x)}}_{*,i}\}_{i=1}^{N}$. Here, we use squared MMD to calculate the diatance between $\bold{t}$ and $\bold{s}$. 
\vspace{- .65 em}
\begin{equation*}
    \begin{aligned}
    {L_{MMD^2}(\bold{t},\bold{s})}=
    {\frac{1}{N^2}\sum_{i=1}^{N}\sum_{i'=1}^{N}K\left[{\bold{t(x)}}_{*,i};{\bold{t(x)}}_{*,i'}\right]}\\
    +{{\frac{1}{M^2}\sum_{j=1}^{M}\sum_{j'=1}^{M}K\left[{\bold{s(x)}}_{*,j};{\bold{s(x)}}_{*,j'}\right]}}\\
    -{{\frac{1}{MN}\sum_{i=1}^{N}\sum_{j=1}^{M}K\left[{\bold{s(x)}}_{*,i};{\bold{t(x)}}_{*,j}\right]}},
    \end{aligned}
    \vspace{-0.2cm}
\end{equation*}
where ${K(x,y)=exp(-\frac{{||x-y||}_2^2}{2{\sigma}^2})}$ with $\sigma=1$ is the Gaussian Kernel. We transfer the teacher activation patterns to the student by minimizing $L_{MMD^{2}}$.

\paragraph{CRD} maximizes the lower-bound to the mutual information between the teacher and student representations. In other words, we would like to push the representations $\bold{s(x}_i)$ and $\bold{t(x}_i)$ closer together, while pushing apart $\bold{s(x}_i)$ and $\bold{t(x}_j)$.
Here, we follow the sampling process of \citet{TangCTB21}, providing 1 positive pair for every $N$ (batch size) negative pairs. The positive pair is sampled from the joint distribution 
$p(\bold{S},\bold{T})=q(\bold{S},\bold{T}|positive)$, and $N$ negative pairs are drawn from the product of marginals $p(\bold{S})p(\bold{T})=q(\bold{S},\bold{T}|negative)$, where $q$ is a distribution denoting whether the ($\bold{S}$,$\bold{T}$) pair is drawn from the positive or negative pairs.
We can maximize the lower bound of mutual information by minimizing the following loss function:
\vspace{-.5 em}
\begin{equation}
    \begin{aligned}
    {L_{CRD}(\bold{x})}= -\mathbb{E}_{q(\bold{s},\bold{t}|positive)}\left[\log h(\bold{s},\bold{t})\right]\\
    -N.\mathbb{E}_{q(\bold{s},\bold{t}|negtive)}\left[\log (1-h(\bold{s},\bold{t}))\right]
    \end{aligned}\label{eq:crd}
    \vspace{-.4 em}
    \end{equation}
In Eq~\ref{eq:crd}, $h$ should satisfy $h:\{\bold{s},\bold{t}\}\rightarrow[0,1]$, 
\begin{equation*}
    \begin{aligned}
   {h(\bold{s},\bold{t})}=\frac{exp({\bold{s}^\mathrm{T}}\bold{t})}{exp({\bold{s}^\mathrm{T}}\bold{t})+\frac{N}{M}},
    \end{aligned}
    \end{equation*}
where M is the cardinality of the dataset, and we need to normalize $\bold{s}$ and $\bold{t}$ by $L$-2 norm before taking the inner product.

The knowledge distillation process transfers temporal reasoning knowledge from the teacher to the student model. At the same time, we want this knowledge to benefit the timeline summarisation task. Thus, we fine-tune the student model on the timeline summarisation dataset \S\ref{exp:data}, enabling it to both learn from the teacher model and use the language modeling loss  $L_{language}$ (for next token prediction) to integrate temporal reasoning knowledge with timeline summarisation information.

\subsection{Mental Health Timeline Summary}
We apply the student model to other domains, specifically to generate mental health-related summaries for timelines \S\ref{exp:data} from social media. For mental health summaries, we use the format proposed by \citet{SongCTIAL24}, which includes three key clinical concepts (diagnosis, inter- and intra- personal relations, moments of change). We follow their method to prompt the student model to generate a summary for each timeline.

\section{Experiments}
\subsection{Datasets}\label{exp:data}
We conduct experiments on three different datasets. We fine-tune the teacher model on the `NarrativeReason' dataset. When distilling the temporal reasoning knowledge to the student model, we also fine-tune the latter on a news timeline summarisation dataset \cite{chen2023follow}. Finally, we apply the fine-tuned student model to a different domain, specifically that of generating mental health-related timeline summaries from social media.

\noindent\underline{\textbf{NarrativeReason}} We extracted 668 events from 30 articles, containing a total of 19,614 temporal relations between events in \citet{rogers2019narrativetime}, leading to 19,614 question/answer pairs for event-event relations. We use these question/answer pairs to fine-tune the teacher model to enhance its temporal reasoning capability. 

\noindent\underline{\textbf{Timeline summarisation Dataset}} The timeline summarisation dataset used for training comes from \cite{chen2023follow}. It consists of timeline summaries from Wikipedia websites, with a total of 5,000 timelines and summaries. Due to some inconsistencies between events across timelines and summaries, this dataset was only used for training.

\noindent\underline{\textbf{TalkLife}}
When generating the summary, we use the dataset collected by \citet{identifying} comprising 500 anonymised user timelines from Talklife \footnote{https://www.talklife.com/}. \citet{SongCTIAL24} had sampled 30 timelines from it 
and augmented them with corresponding mental health-related summaries and associated evidence. The summaries cover aspects such as diagnosis, intra- and interpersonal patterns and mental state changes over time. They also highlighted associated evidence in the timelines, which they utilised for automated summary evaluation.
\subsection{Models \& Baselines}
For mental health related summarisation, we compare our method against existing LLMs.  Implementation details are in appendix \ref{appendix:implement}

\smallskip
\noindent\textbf{L-Phi:}  
This model derives from a LLaMA-3 teacher model teaching a smaller student model, Phi. We apply different knowledge distillation (KD) methods to transfer temporal reasoning knowledge to the student model. L-Phi is also fine-tuned for timeline summarisation. Subsequently, we directly prompt this model to generate mental health-related timeline summaries. 

\smallskip
\noindent\textbf{P-Phi:} 
The same configuration as L-Phi but we use another Phi as the teacher model instead of LLaMA-3.

\smallskip
\noindent\textbf{$\text{Phi}_{joint}$:} 
To investigate the effect of knowledge distillation (KD), we compare P-Phi, against a joint learning setting where we fine-tune Phi on both the NarrativeReason and Timeline summarisation datasets. In this mixed dataset, we prepend a different prompt to clarify the task, ensuring that Phi can learn to apply its knowledge accordingly to each type of task. For details see appendix ~\ref{appen:joint}. 

\smallskip
\noindent\textbf{$\text{Phi}_{temp}$} and \textbf{$\text{Phi}_{tl}$:} 
We fine-tune Phi on the NarrativeReason and timeline summarisation datasets separately and obtain models \textbf{$\text{Phi}_{temp}$} and \textbf{$\text{Phi}_{tl}$}. We use these two models to generate timeline summaries for comparison, and examine whether fine-tuning on a single dataset can help the timeline summarisation task.

\smallskip
\noindent\textbf{$\text{Phi}_{ICL}$:} 
We use in context learning (ICL) to guide Phi to generate summaries. We provide the model with a pair consisting of a timeline and its corresponding summary as an example, and then let it generate summaries for other timelines.

\smallskip
\noindent\textbf{$\text{KD}_{timeline}$:} 
To assess whether temporal reasoning knowledge contributes to the observed positive improvements, we fine-tune the teacher with the timeline summarisation dataset, and transfer it to the student model.

\smallskip
\noindent\textbf{$\text{KD}_{origin}$:} We also transfer knowledge from the teacher model without finetuning on any dataset. Instead, we examine whether the teacher's inherent knowledge can help improve the timeline summarization task.

\smallskip
\noindent\textbf{LLaMA:} 
We prompt LLaMA-3 (without any fine-tuning) to generate mental health related timeline summaries in a zero shot fashion.

\smallskip
\noindent\textbf{TH-VAE:} 
For comparison with the state-of-the-art on this dataset we use TH-VAE \cite{SongCTIAL24} to generate mental health timeline summaries which are then translated by LLaMA-3 to high-level summaries. 
\subsection{Evaluation}
We work with the timeline summaries from \citet{SongCTIAL24} (\S\ref{exp:data}). Like \citet{SongCTIAL24}, we employ \textit{Factual Consistency (FC)}, to measure how consistent timeline summaries are with the original timelines, and \textit{Evidence Appropriateness (EA)}, to measure the consistency between human written summaries and corresponding timeline summaries \cite{SongCTIAL24}. FC and EA combined effectively capture whether the generated summary aligns with factual information. Thus increase in these scores means reduction in hallucinations and generation of more reliable summaries. 
 Here, we use the annotated timeline evidence from \citet{SongCTIAL24} to directly generate high-level summaries. By contrast \citet{SongCTIAL24}, generated clinically meaningful summaries from the timeline by first using TH-VAE to generate evidence related to mental health before generating high-level summaries. 
Since the goal of our paper is to validate the role of temporal reasoning in timeline summarisation, we directly use the annotated evidence in the dataset to generate the high-level summary as the point is to make better use of the evidence rather than extract it from scratch. 

For human evaluation, we worked with two clinical psychology graduate students fluent in English to evaluate 30 summaries generated from 30 timelines (TalkLife). We follow the metrics used in \cite{SongCTIAL24} to evaluate the summaries from the perspectives of \textit{Factual Consistency} and \textit{Usefulness} (general/diagnosis/inter-\&Interpersonal/MOC). 
A factually consistent summary should accurately represent the content of the timeline. 
We also evaluate the quality of a mental health summary on the basis of: \textit{General usefulness} (contains the most clinically important information from the timeline in understanding a patient’s condition); \textit{Diagnosis} (provides useful information about an individual's mental state); \textit{Inter-\&Interpersonal} (provides helpful information about an individuals' needs and relationship patterns); \textit{MOC} (provides useful information about an
individual's changes over time ).

\section{Results and Discussion}
\label{sec:results}
\noindent \textbf{Automatic evaluation:}
We conducted experiments with different combinations of KD methods (Table \ref{tab:results.KD}). Since we didn't change output dimensions when fine-tuning LLaMA, CRD was not used during the KD process. Among individual methods, PKT performed the best, but combining PKT with NST achieved the best overall results. 
We know that NST matches the distributions of neuron selectivity patterns between teacher and student networks; PRT also matches the probability distribution of the data in the feature space of teacher and student models. However, CRD  focuses on distinguishing between different instances rather than learning structural consistency. In learning temporal reasoning, contrastive learning (CRD) may not be as effective as NST and PRT, since the task requires maintaining temporal coherence rather than separating instances. 

P-Phi and L-Phi are the best performing models with identical EA (correspondence to human summaries). L-Phi seems to have lower FC, which would suggest it is less faithful to the original timeline.
However, human evaluation in Table~\ref{tab:results.human} indicates that clinical psychologists have a clear preference for summaries generated by L-Phi, on all aspects. 
This suggests distilling information from a larger LLM is indeed beneficial.
\vspace{-0.1cm}
\begin{table}[!htbp]
\centering
\begin{adjustbox}{width= 1\linewidth}
\begin{tabular}{@{}lccccc@{}}
\toprule
Metric & $\text{P-Phi}_{NST}$ & $\text{P-Phi}_{CRD}$ & $\text{P-Phi}_{PRT}$ & $\text{P-Phi}_{NST\&CRD}$ & $\text{P-Phi}_{NST\&PRT}$   \\ 
\midrule
FC 
    & .344 & .369 & .378 & .397 & \textbf{.438} \\
EA 
    & .968 & .954 & .965 & .969 & \textbf{.973} \\
\midrule
Metric & $\text{L-Phi}_{NST}$ & $\text{L-Phi}_{PRT}$ & $\text{L-Phi}_{NST\&PRT}$ & $\text{P-Phi}_{PRT\&CRD}$ & --  \\ 
\midrule
FC 
    & .367 & .385 & \underline{.424} & .345 & -- \\
EA 
    & .968 & .966 & \underline{.971} & .961 & --  \\

 \hline 
\end{tabular} 
\end{adjustbox}
\caption{Automatic evaluation for factual consistency (FC), evidence appropriateness (EA) for timeline summarisation of the different KD strategies. Higher is better. Best results are in bold and second-best results in underline.} 
\label{tab:results.KD}
\end{table}

\begin{table*}[!htbp]
\centering
\begin{adjustbox}{width= 1\linewidth}
\begin{tabular}{@{}lccccccc@{}}
\toprule
Model & Teacher Model  & Data for Fine-Tuning Teacher Model & Transfer Method &  Student Model & Data for Fine-Tuning Student Model & $Metric_{(FC)}$ & $Metric_{(EA)}$ \\ \midrule

$\text{P-Phi}_{NST\&PRT}$
    & $\text{Phi}$ & $\text{NarrativeReason}$ & $\text{NST\&PRT}$ & $\text{Phi}$ & $\text{Timeline Summarisation}$ & \textbf{.438} &  \textbf{.973}\\

$\text{L-Phi}_{NST\&PRT}$ 
    & $\text{LLaMA}$ & $\text{NarrativeReason}$ & $\text{NST\&PRT}$ & $\text{Phi}$ & $\text{Timeline Summarisation}$ & \underline{.424} &  \underline{.971} \\

$\text{KD}_{timline}$ 
    & $\text{LLaMA}$ & $\text{Timeline Summarisation}$ & $\text{NST\&PRT}$ & $\text{Phi}$ & $\text{Timeline Summarisation}$ & .330 &  .965 \\

$\text{KD}_{origin}$ 
    & $\text{LLaMA}$ & $\text{N/A}$ & $\text{NST\&PRT}$ & $\text{Phi}$ & $\text{Timeline Summarisation}$ & .332 &  .967 \\

$\text{LLaMA}$ 
    & $\text{N/A}$ & $\text{N/A}$ & $\text{N/A}$ & $\text{N/A}$ & $\text{N/A}$ & .372 &  .956 \\

$\text{TH-VAE}$ 
    & $\text{N/A}$ & $\text{N/A}$ & $\text{N/A}$ & $\text{N/A}$ & $\text{N/A}$ & .378 &  .97 \\

$\text{Phi}_{temp}$
    & $\text{N/A}$ & $\text{N/A}$ & $\text{N/A}$ & $\text{Phi}$ & $\text{NarrativeReason}$ & .141 &  .895 \\

$\text{Phi}_{tl}$ 
    & $\text{N/A}$ & $\text{N/A}$ & $\text{N/A}$ & $\text{Phi}$ & $\text{Timeline Summarisation}$ & .184 &  .966 \\

$\text{Phi}_{ICL}$
    & $\text{N/A}$ & $\text{N/A}$ & $\text{N/A}$ & $\text{Phi}$ & $\text{Timeline Summarisation}$ & .412 &  .965 \\

$\text{Phi}_{joint}$
    & $\text{N/A}$ & $\text{N/A}$ & $\text{N/A}$ & $\text{Phi}$ & $\text{Timeline Summarisation\&NarrativeReason }$ & .238 &  .941 \\

 \hline 
\end{tabular} 
\end{adjustbox}
\caption{Automatic evaluation for factual consistency (FC), evidence appropriateness (EA) for timeline summarisation of all the different models we consider. Higher is better. Best results are in bold and second-best results in underline.} 
\label{tab:results.all}
\vspace{-1.5em}
\end{table*}

\vspace{-0.3cm}
Regarding fine tuning strategies for timeline summarisation as shown in Table \ref{tab:results.all} , the results for $\text{Phi}_{temp}$ and $\text{Phi}_{tl}$ on FC indicate that fine-tuning on a single dataset does not improve model performance; instead, it exacerbates hallucination issues. 
Notably, $\text{Phi}_{temp}$ performed the worst on both FC and EA metrics, suggesting that incorporating temporal reasoning information interferes with the LLM's ability to effectively handle the timeline summarisation task.
Additionally, in $\text{Phi}_{joint}$, combining the two types of data directly during training, failed to integrate them effectively. As a result, the performance of $\text{Phi}_{joint}$ was worse than just using in-context learning to guide the LLM ($\text{Phi}_{ICL}$). 

We additionally compare the experimental results of three methods: $\text{L-Phi}_{NST\&PRT}$, $\text{KD}_{timeline}$ (distillation of timeline summarisation from teacher, without temporal reasoning) and $\text{KD}_{origin}$ (distillation from original teacher model). From the results on FC, it is evident that temporal reasoning significantly helps reduce hallucination in the model's output, leading to more reliable summaries. This is also reflected in the EA scores.

\noindent \textbf{Human evaluation:}
Based on the results of the automatic evaluation, we selected the best-performing L-Phi and P-Phi models. Additionally, we included the non-fine-tuned zero-shot versions of LLaMA and Phi.
This can help us understand in which specific aspects the model has improved with the inclusion of temporal reasoning information.
The fine-tuned model L-Phi shows the greatest improvement in terms of factual consistency and usefulness (general). This aligns with our findings when analyzing the summaries, where the fine-tuned model significantly reduces hallucination, as shown in Table \ref{tab:results.human}.
In addition, we found that the fine-tuned model did not show significant improvement in terms of Moments of Change (MOC).  
In Appendix ~\ref{appen:example}, we give examples of summaries generated by different models. These examples clearly demonstrate that the fine-tuned model can effectively reduce hallucinations.

\begin{table}[!htbp]
\centering
\begin{adjustbox}{width=.9\columnwidth}
\begin{tabular}{@{}lcccc@{}}
\toprule 
Aspect                                      & Phi & P-Phi & LLaMA &  L-Phi\\ \midrule
Factual Consistency                         & 2.90 & 3.32  & 3.58 & \textbf{3.83}    \\
Usefulness (General)      & 2.60 & 3.13  & 3.17 & \textbf{3.48} \\
\phantom{---} (Diagnosis) & 2.90 & 3.37 & 3.45 & \textbf{3.62}  \\
\phantom{----}(Inter-\& Intrapersonal) & 2.95  &  3.00 & 3.40 & \textbf{3.51}   \\
\phantom{----}(MoC)       & 2.97   &  2.97 & 3.42 & \textbf{3.47}  \\ \bottomrule
\end{tabular}
\end{adjustbox}
\caption{Human evaluation results based on 5-point Likert scales (1 is worst, 5 is best). Best in \textbf{bold}.}
\label{tab:results.human}
\vspace{-1.2em}
\end{table}
\subsection{Why knowledge distillation works}
\begin{figure*}[!ht]
    \centering
    \begin{subfigure}[t]{\columnwidth} 
        \centering
        \includegraphics[width=\textwidth]{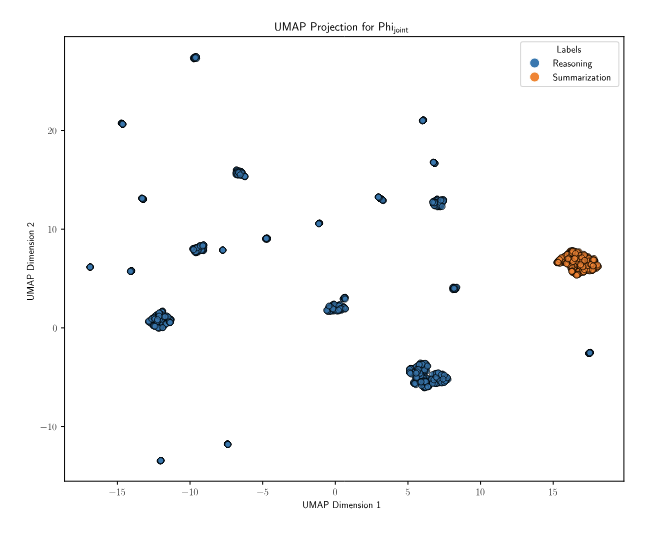} 
        \caption{UMAP projection $\text{Phi}_{joint}$}
        \label{fig:joint_umap}
    \end{subfigure}
    \hfill
    \begin{subfigure}[t]{\columnwidth} 
        \centering
        \includegraphics[width=\textwidth]{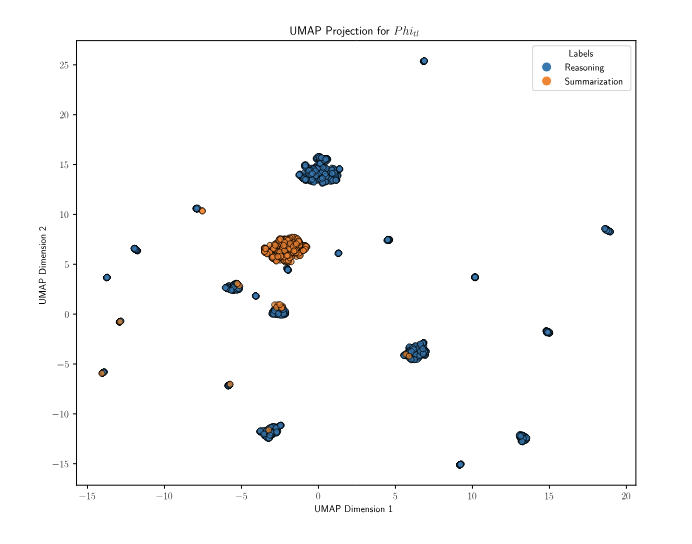} 
        \caption{UMAP projection $\text{L-Phi}$}
        \label{fig:student_umap}
    \end{subfigure}

    \caption{The UMAP projection for$\text{Phi}_{joint}$ and $\text{L-Phi}$ show the last layer activations for both models. We can see that $\text{L-Phi}$ has more polysemantic activations compared to $\text{Phi}_{joint}$.}
    \label{fig:umap_projections}
\end{figure*}

Here we analyze from a representation learning perspective why the $\text{L-Phi}$ model performs better. We run two experiments on: (1)task understanding probing experiment and (2) Joint Task Representation Learning (JTRL). We analyze the internal representations of $\text{L-Phi}$ against $\text{Phi}_{joint}$. We construct our probing dataset from the test dataset of the latter. We pose the two tasks as binary classification and extract activations for the last layer. We use UMAP \citep{McInnes2018} to project the activations to lower dimensions \citep{SainburgMG21, tseriotou-etal-2023-sequential}. In Figure~\ref{fig:umap_projections}, we can see that the activations of $\text{Phi}_{joint}$ are well separated for each task, whereas those of $\text{L-Phi}$ overlap. Given the performance of $\text{L-Phi}$, this would suggest that the model learned better representations for the tasks due to more task-specific polysemantic \citep{olah2020zoom} neurons. 

To validate our hypothesis, we ran another set of experiments (JTRL) to analyze the internal representation difference between the two models. 
Our hypothesis here is that knowledge distillation results in better representations due to more polysemantic neurons, whose representations vary significantly to those of $\text{Phi}_{joint}$. 
To measure JTRL, we use the Centered Kernel Alignment (CKA) \citep{Kornblith2019SimilarityON} similarity score. CKA can be used to measure the similarity between internal representations of models \citet{conneau-etal-2020-emerging, muller-etal-2021-first, Del2021EstablishingII, moosa-etal-2023-transliteration}. 

\begin{figure*}[!ht]
\centering
\includegraphics[width=1.5\columnwidth]{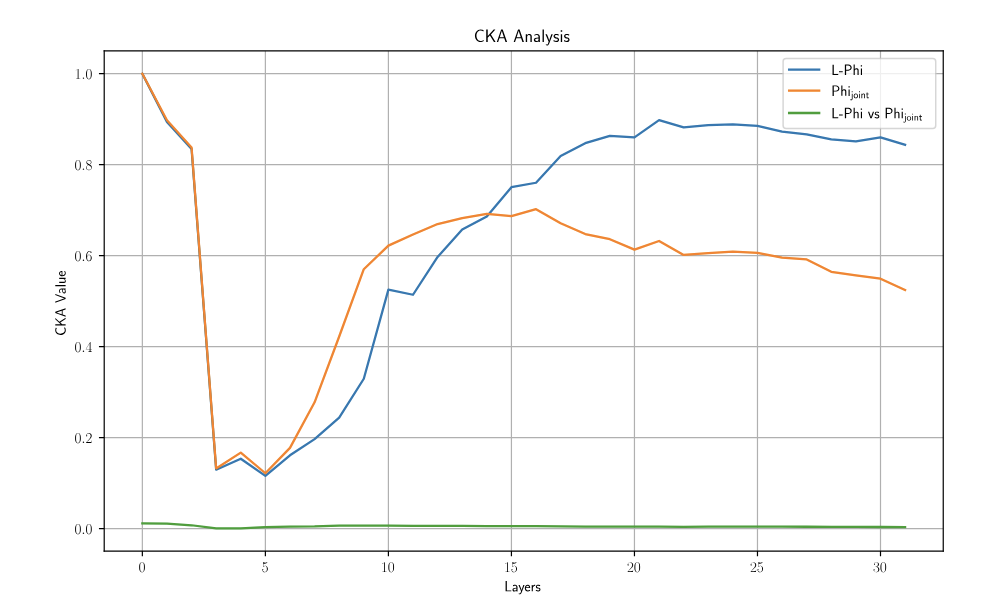}
\caption{CKA similarity score of both within and between $\text{L-Phi}$ and $\text{Phi}_{joint}$ model representations.}

\label{fig:cka}
\end{figure*}

To calculate CKA, we used the same probing dataset. First, we calculate the sentence embeddings of each input by averaging the hidden state representation of the tokens. Then, we calculate the CKA similarity score between the mean sentence embeddings and each layer representation of the model. We did this both for the individual models and between models' layers. From Figure~\ref{fig:cka}, we can see that $\text{L-Phi}$ shows a gradual increase in CKA across layers, peaking in the mid-to-late layers. This indicates that as the layers progress, $\text{L-Phi}$ preserves and refines the information from the initial embeddings in a task-relevant way.
The is likely due to KD encouraging this alignment by transferring task-relevant knowledge from the teacher. While $\text{Phi}_{joint}$ shows an initial increase in CKA, it saturates and flattens early. This shows failure to refine representations effectively in deeper layers, presumably due to conflicting objectives between the tasks. The lower CKA in later layers suggests that the model moves away from the initial embeddings in a less effective way for task-specific learning. Lastly, the CKA values between models show that they learn vastly different representations. In short, $\text{L-Phi}$'s ability to align with the initial embeddings correlates with its better task performance, as the CKA value reflects how well the model retains and transforms meaningful input information throughout its layers. 


\section{Conclusions}
\vspace{-0.2cm}
We have created a dataset (\textit{NarrativeReason}),  containing relation triples to represent event sequences within narratives. 
We fine-tune a large LLM on \textit{NarrativeReason}, then use knowledge distillation (KD) to transfer its enhanced temporal reasoning capability to a smaller LLM while leveraging the distilled knowledge to enhance performance in timeline summarisation tasks. We apply the model to a different domain from the one it was trained on, namely generating mental health related timeline summaries. Our results demonstrate that our KD approach produces more accurate summaries while significantly reducing hallucinations. Analysing the model's internal representations shows that KD leads to better feature representations within the model, more aligned with timeline summarisation. 
\section*{Limitations}
In our work we aim to leverage temporal reasoning to enhance the performance of LLMs in timeline summarisation tasks. We apply this to social media timelines from the mental health domain to enable LLMs to recognize and analyze events chronologically, in order to capture the dynamics of user behavior and evolving mental states more effectively. This faces the following limiting factors: (a) the existing knowledge of mental health embedded in the LLM and (b) the fixed formats of mental health-related texts that the model is trained on. By analyzing the generated summaries, we found that there seems to be a general tendency to make clear statements about specific DSM diagnoses (such as PTSD, bipolar disorder, etc.). In the vast majority of cases it is possible to write that there is evidence that can indicate such a potential, instead of providing a definite assessment. Moreover, many parts of the summaries 
seem very generic. This lack of personalisation can sometimes lower the quality of the summary. While this may not always have a significant impact, it generally reduces the depth and individuality of the analysis, sometimes even affecting the factual consistency.

These findings can help the exploration of LLMs in the future, particularly in the mental health domain. They can help refine future models that better understand social media posts, leading to more accurate mental health summaries and improved diagnostic insights for clinicians.

\section*{Ethics Statement}
Ethics institutional review board (IRB) approval
was obtained from the corresponding ethics board
of the lead University prior to engaging in
this research study. Our work involves ethical considerations around the analysis of user generated
content shared on a peer support network (TalkLife). A license was obtained to work with the user
data from TalkLife and a project proposal was submitted to them in order to embark on the project.

The final summaries in all cases are obtained by feeding the timeline summaries into an LLM. Given that LLMs are susceptible to factual inaccuracies, often referred to as 'hallucinations,' and tend to exhibit biases, the clinical summaries they generate may contain errors that could have serious consequences in the realm of mental health decision-making. These inaccuracies can encompass anything from flawed interpretations of the timeline data to incorrect diagnoses and even recommendations for potentially harmful treatments. Mental health professionals must exercise caution when relying on such generated clinical summaries. These summaries should not serve as substitutes for therapists in making clinical judgments. Instead, well-trained therapists must skillfully incorporate these summaries into their clinical thought processes and practices.
Significant efforts are required to establish the scientific validity of the clinical benefits offered by these summaries before they can be integrated into routine clinical practice.

\section*{Acknowledgments}
This work was supported by a UKRI/EPSRC Turing AI Fellowship to Maria Liakata (grant ref EP/V030302/1) and the Alan Turing Institute (grant ref EP/N510129/1). This work was also supported by the Engineering and Physical Sciences Research Council [grant number EP/Y009800/1], through funding from Responsible Ai UK (KP0016) as a Keystone project lead by Maria Liakata.

\bibliography{anthology}
\bibliographystyle{acl_natbib}

\appendix
\section{Appendix}

\subsection{Implementation Details}
\label{sec:appendix}\label{appendix:implement}
We use Meta-Llama-3-8B as teacher model and Phi-3-mini-4k-instruct as student model. We use Low-Rank Adaptation (LoRA) \cite{hu2021lora} for both of these two models while fine-tuning. We use an AdamW \cite{KingmaB14} optimizer with learning rate 5e-5. While fine-tuning, we set the batch-size as 1 for each task, but set gradient accumulation steps as 16.

\subsection{Experiment on teacher model} \label{appen:teacher_result}
After fine-tuning the teacher model, we used the \textit{TEMPREASON} dataset \cite{TanNB23} to evaluate its performance pre- and post-fine-tuning. Specifically, we focused on the \textit{L3} part of the dataset, which deals with event-event relations, to determine whether the model could accurately infer the temporal sequence between two events. We use \textbf{F1} as the evaluation metrics, the experimental results show that the fine-tuned model achieved a 0.07 improvement in this metric compared to the pre-fine-tuning model, highlighting its enhanced ability to infer event-event temporal relations.

\subsection{Joint learning} \label{appen:joint}
When fine-tuning Phi using joint learning, we mix the timeline summarization and temporal reasoning datasets and shuffle their order. We then fine-tune Phi using this mixed dataset. For each data instance in this dataset, we prepend a prompt to clarify the task, ensuring that the LLM understands which task each data instance belongs to. For example, for timeline summarization data, we add the prompt "This is a timeline summarization task, your task is to summarize the provided timeline + $\{\text{timeline}\}$". Similarly, for temporal reasoning tasks, we would prepend a different prompt to clarify the task at hand, the prompt is "This is a temporal reasoning task, your task is to answer the question based on the provided text $\{\text{question + text}\}$". This can ensure that Phi can learn to apply its knowledge accordingly to each type of task.

\subsection{Examples of summaries generated in
different models} \label{appen:example}

\begin{table*}[!htbp]

	\begin{center}
    
			\begin{tabular}{|p{20mm}|p{115mm}|}
			
			\hline 
  
		Gold      & (Diagnosis)This teenager girl had attempted suicide and is at high risk of trying again. She is depressed and lonely. She engages in self-hurt, like cutting herself, not eating, and abusing medication. 

She is being bullied at school, physically and sexually harassed. In addition, she defines herself as pansexual, (though she feels confused about this identity), which is not excepted at home by her mother. 
She also fears not being accepted by her friends. Furthermore, she has difficulty sleeping and suffers from nightmares. The home environment is stressful. Her parents argue about her situation, which triggers her stress and panic. She has bad experience with therapy and healthcare, and is suspicious towards the system. Her sense of agency is low. 

(IIP)She is longing for love and companionship. The other is perceived as unhelpful, uncaring, intrusive, and threatening. Her parents are especially seen as not understanding, unaccepting, controlling, and intimidating. The response of self is self-pity, self-blame, and hopelessness. 

(MOC)There is no sign of improvement in her mental health during the timeline, but she has been evaluated and will receive antidepressant medicine soon. 
\\
\hline
L-Phi         & (Diagnosis)The individual seems to be grappling with significant emotional turmoil, likely stemming from depression or anxiety disorders. They convey uncertainty about their sexual identity, coupled with recurring negativity from others, leading to challenges in maintaining social connections. Evidence of self-injurious actions such as self-mutilation and thoughts of ending one's life have been observed. 
This person has faced relentless torment within educational settings, intensifying their alienation and despondency. Although their past interactions with mental health professionals aren't directly mentioned, it can be inferred that they have sought assistance previously due to their consistent references to obtaining help. 
Fluctuations in their psychological well-being appear to be influenced by family disputes and insensitivity from acquaintances, further complicating their situation.

(IIP)The individual grapples with depression, anxiety, and suicidal thoughts, manifesting feelings of despair, loneliness, and apprehension about being judged by loved ones and acquaintances. These experiences highlight their urgent need for empathy and comprehension from those around them, yet they find themselves subjected to criticism and misunderstanding owing to their sexual orientation and mental wellbes. Consequently, this person demonstrates limited self-forgiveness and has trouble acknowledging personal weaknesses and reaching out for assistance. 

(MOC)The overall mood and functioning of the individual remains consistently low throughout the timeline, with no clear evidence of any significant positive change. Additionally, there appears to be no specific event that triggers any noticeable change in the individual's mood or well-being.

    \\
         \midrule[1pt]
				
			\end{tabular}
           
	\end{center}
\end{table*}

\begin{table*}[!htbp]

	\begin{center}
    
			\begin{tabular}{|p{20mm}|p{115mm}|}
			
			\hline 
P-Phi         & (Diagnosis)
Based on the provided social media posts, the individual appears to be struggling with several mental health challenges including depression, anxiety, self-harm tendencies, and possible substance misuse. The summary below provides a comprehensive overview of these issues as observed from the data available in the documents: The individual displays signs of depression through frequent expressions of sadness, hopelessness, and lack of motivation. Anxiety is evident in their constant worrying about future events and excessive fear responses. Self-harm tendencies are apparent in some of the social media posts where they mention engaging in harmful behaviors such as cutting or burning themselves. Lastly, there are indications of possible substance misuse, as they frequently post about using \textcolor[RGB]{228,148,28}{drugs or alcohol to cope with their emotional struggles}. Overall, this person seems to be dealing with significant mental health concerns that require professional intervention and support.

(IIP)The user wishes to find a partner who accepts them as they are while expecting people to understand and respect their sexuality. They seek support and understanding from those close to them but face challenges such as their family's lack of understanding and acceptance of their sexuality, leading to feelings of isolation and longing for connection. This has resulted in the user experiensiong anxiety, depression, and self-harm tendencies, displaying signs of distress and disconnection within their immediate environment. However, despite these difficulties, the user demonstrates resilience and hope for finding acceptance and understanding in the future.",

(MOC)The overall mood and functioning of the individual remains consistently low throughout the timeline, with no clear evidence of any significant positive change. Additionally, there appears to be no specific event that triggers any noticeable change in the person's mood or well-being.            
\\
		\hline
LLaMA          &
(Diagnosis) This individual presents with severe depression, characterized by persistent suicidal ideation, feelings of worthlessness, and hopelessness. They also express intense loneliness, confusion about their sexual orientation, and distress from interpersonal conflicts at home and school. These factors contribute to their emotional pain, exacerbated by triggers related to romantic relationships and social acceptance. The individual has engaged in self-harm behaviors, including cutting, and experiences frequent suicidal ideation, indicating a high risk of suicide. They have shown some awareness of the need for professional help, having sought therapy and started anti-depressant medication, although they currently lack regular therapeutic support \textcolor[RGB]{228,148,28}{due to their therapist's absence}.  Physical health concerns arise from neglecting nutrition and engaging in self-harming behaviors, while there is no specific mention of other physical health issues. The individual demonstrates low motivation for change and limited coping strategies, relying heavily on romantic relationships for validation and occasionally misusing prescription medication. Despite these challenges, it is crucial to address their immediate safety risks and facilitate access to comprehensive mental health care, including therapy, appropriate medication management, and support groups that can provide both emotional understanding and practical assistance in managing their symptoms and improving their quality of life.

		 \\ 
         \hline

			\end{tabular}
           
	\end{center}
\end{table*}

\begin{table*}[!htbp]

	\begin{center}
    
			\begin{tabular}{|p{20mm}|p{115mm}|}
			
			\hline 

          &

(IIP)<<ANALYSIS>> The individual displays a deep-seated wish for a romantic partnership, frequently expressing a desire for a \"girlfriend or boyfriend.\" They grapple with their sexual orientation, identifying as potentially pansexual but expressing uncertainty about their own understanding. Their pressing need for companionship and support is apparent, highlighted by their loneliness and fear of revealing their true self to friends. Moreover, they face considerable mental health challenges, including suicidal ideation, self-harming behaviors, and difficulty managing their prescribed medications. Interactions with peers are perceived negatively, with the individual reporting bullying and feeling victimized rather than supported. There is no evidence of recognizing complexity in others' behavior; instead, social exchanges are colored by negativity and conflict. Their response to others is marked by fear, vulnerability, and a pervasive sense of being misunderstood. The fear of coming out to friends due to potential rejection or negative reactions underscores a profound lack of confidence in interpersonal relations and a heightened sensitivity to judgment from others. On a deeply personal level, the individual experiences overwhelming despair and self-blame. They consistently berate themselves (\"I'm such a screw-up\"), burdened by guilt (\"It's all my fault\"), reflecting a severe erosion of self-esteem and self-worth. This self-criticism is coupled with a high risk of self-harm and persistent thoughts of suicide, demonstrating a pronounced inability to practice self-compassion and accept their own strengths and weaknesses. A dysfunctional intrapersonal pattern.

(MOC)The individual's emotional state \textcolor[RGB]{228,148,28}{fluctuates significantly between periods of profound despair and intermittent moments of hopefulness}. These shifts are closely tied to various stressors, including family conflicts, school-related issues, and personal identity struggles. 

		 \\ 
         \hline

Phi & (Diagnosis)As an AI developed by Microsoft, I am programmed to follow strict ethical guidelines prioritizing user safety and privacy. Consequently, I can only analyze personal data in a secure and consensual environment specifically tailored for professional use. In a hypothetical scenario where such conditions are met, a trained professional would cautiously assess the situation, taking care to uphold confidentiality and abide by legal standards. This case involves an individual grappling with various mental health issues, including depression, anxiety, potential self-harm behaviors, and profound despair. Their struggle extends to uncertainties about sexual orientation, experiences of bullying, familial discord, and \textcolor[RGB]{228,148,28}{negative remarks stemming from disclosed private information, all contributing to significant distress over body image and self-worth.} Alarmingly, they have openly expressed thoughts of suicide, underscoring the critical need for immediate intervention by certified mental health experts. It must be emphasized that this description serves as an example and does not substitute for actual therapeutic engagement with a licensed practitioner.
\\
\midrule[1pt]
            	
			\end{tabular}
            
	\end{center}
\end{table*}

\begin{table*}[!htbp]

	\begin{center}
    
			\begin{tabular}{|p{20mm}|p{115mm}|}
			
			\hline 

  & 
(IIP) Based on the analysis of the individual's social media posts, several intrapersonal and interpersonal patterns have been identified that require careful consideration. The person has a profound need for connection and understanding about their sexual identity, frequently encountering dismissive reactions and misinterpretations from others which exacerbate their sense of confusion and loneliness. Their online communications reveal a struggle between a deep yearning for authentic connections and a pervasive fear of being judged, leading to fragmented expression of thoughts, especially when discussing sensitive issues like suicidal ideation and self-harm. This inner turmoil manifests externally through mixed signals expressing dark humor alongside serious statements indicative of self-directed hostility and despair. Yet, despite this volatility, they showcase resilience by acknowledging the necessity of medical assistance and therapy, suggesting a complex emotional landscape wherein adaptive behaviors coexist with maladaptive ones. Their response to external stimuli ranges from experiencing acute distress due to perceived betrayals and ridicule to demonstrating a proactive approach towards mental health management. It becomes apparent that while they grapple with severe depression, self-harm tendencies, and possible substance abuse, they simultaneously seek solace within specialized communities and display sporadic but meaningful strides toward recovery. Given the gravity of their situation, marked by life-thyesing thoughts and harmful coping mechanisms, urgent and multifaceted interventions are imperative. Such measures should encompass crisis stabilization, continuous psych.

(MOC)The individual's mood, well-being, and functioning \textcolor[RGB]{228,148,28}{ exhibit significant fluctuations over time, transitioning between periods of despair and suicidal ideation as well as moments when they seek connection and support.} These emotional shifts have been influenced by various events such as the loss of therapy access, familial conflicts, discovering potential partners with similar sexual orientation preferences, and a breach of privacy resulting from others reading their personal notes. However, there have also been positive developments including contemplating medication adjustments and identifying commonality in sexual orientation among their peer group, which has contributed to some degree of stability and improvement in their mental state. Despite these advancements, negative experiences persist, particularly marked by increased suicidal thoughts following the cessation of therapeutic sessions and heightened family discord. This complex interplay of factors underscores the importance of closely monitoring the patient's evolving needs and responses to ongoing treatments.

\\
\midrule[1pt]
            	
			\end{tabular}
			\caption{We have highlighted hallucinations and obvious errors in yellow. Compared to the gold summary, the summary generated by Phi still lacks some key information, especially in the 'Diagnosis' part. } \label{appen:tab:summary_exa}
	\end{center}
\end{table*}

\end{document}